\pdfoutput=1

\documentclass[11pt]{article}

\usepackage[]{acl}

\usepackage{times}
\usepackage{latexsym}

\usepackage[T1]{fontenc}

\usepackage[utf8]{inputenc}

\usepackage{microtype}

\usepackage{amsmath}
\usepackage{enumitem}
\usepackage{color}
\usepackage[font={footnotesize}]{subcaption}
\usepackage{multirow}
\usepackage{tabularx}
\newcolumntype{Y}{>{\arraybackslash}X}
\usepackage{booktabs}
\usepackage{array}
\usepackage[font=small,belowskip=-5pt,aboveskip=3pt]{caption}
\usepackage{soul} 
\usepackage{graphicx}

\font\ssf=cmssdc10
\def\Black{\pdfliteral{0 g}}

\def\White{\pdfliteral{1 g}}
\def\Gray{\pdfliteral{.945 .941 .941 rg}}
\def\Green{\pdfliteral{.404 .722 .408 rg}}
\def\bcolor#1{\if r#1\Green \else \Gray \fi}
\def\fcolor#1{\if r#1\White \else \Black \fi}

\def\bptopt#1{.9963 0 0 .9963 0 0 cm 9 0 0 #117 0 3.5 cm }
\def\circle{.5 0 m .5 .276 .276 .5 0 .5 c -.276 .5 -.5 .276 -.5 0 c
           -.5 -.276 -.276 -.5 0 -.5 c .276 -.5 .5 -.276 .5 0 c }
\def\hcircle{.5 .276 .276 .5 0 .5 c -.276 .5 -.5 .276 -.5 0 c }
\def\fullround{\pdfliteral{q \bptopt+ \circle f Q}}
\def\halfround{%
   \ifcase\count255
       \pdfliteral{q \bptopt+ -.5 -.5 m -.5 .5 l .5 .5 l .5 -.5 l h f Q}%
   \or \pdfliteral{q \bptopt+ -.5 -.5 m .5 -.5 l .5 0 l \hcircle  h f Q}%
   \or \pdfliteral{q \bptopt- -.5 -.5 m .5 -.5 l .5 0 l \hcircle  h f Q}%
   \or \fullround \fi
}
\def\xround#1#2{\ifdim\ht1>12pt
   \dimen0=\ht1 \advance\dimen0 by-12pt
   \hbox to0pt{\hss\vbox{\xroundA#1#2\kern\dimen0 \xroundA#1#2}%
      \rlap{\kern-8.5pt\raise4.5pt\vbox{\hrule width17pt height\dimen0}}\hss}%
   \else \xroundA#1#2\fi}
\def\xroundA#1#2{\if #1#2\halfround \else \fullround\fi}

\long\def\messenger#1#2{\medskip\setbox0=\vbox{\penalty13
   \widowpenalty=0 \clubpenalty=0 \interlinepenalty=0
   \everypar={\setbox0=\lastbox\endgraf\vbox\bgroup
      \everypar={\vrule height12pt width0pt}}
   \def\par{\ifhmode\endgraf\egroup\fi}
   \def\\{\unskip\break}\rightskip=0pt plus 1fil\parindent=0pt \ssf #2\par
   \everypar={}\let\par=\endgraf \setbox1=\box2
   \loop \setbox1=\lastbox
       \unless\ifvoid1
          \unskip\unskip 
          \count255=\ifvoid2 \ifnum\lastpenalty=13 3\else 2\fi 
                    \else    \ifnum\lastpenalty=13 1\else 0\fi \fi
          \unpenalty
          \ifdim\ht1>12pt \else
             \setbox0=\vbox{\unvbox1 \lastbox \setbox1=\lastbox 
                      \global\setbox1=\hbox{\unhbox1\unskip\unskip}}
          \fi
          \setbox1=\hbox to\hsize{\hss\bcolor#1\xround l#1%
              \rlap{\vrule height\ht1 depth5pt width\wd1 \xround r#1}%
              \fcolor#1\box1 \Black \if l#1\hfill\fi}
          \global\setbox2=\vbox{\box1 \vskip-3pt \unvbox2}
   \repeat
   }
   \unvbox2
   \medskip
}

\setul{0.5ex}{0.3ex}
\setulcolor{red}
\title{Context-Aware Abbreviation Expansion Using Large Language Models}

\author{
Shanqing Cai$^*$ \ ~ Subhashini Venugopalan\thanks{ ~ equal contribution} \ ~ Katrin Tomanek \\
{\bf Ajit Narayanan} \ ~  {\bf Meredith Ringel Morris} \ ~  {\bf Michael P. Brenner} \\
  Google Research \\
  \texttt{\{cais,vsubhashini\}@google.com}

  }

\begin{document}
\maketitle

\begin{abstract}
Motivated by the need for accelerating text entry in augmentative and alternative communication (AAC) for people with severe motor impairments, we propose a paradigm in which phrases are abbreviated aggressively as primarily word-initial letters. Our approach is to expand the abbreviations into full-phrase options by leveraging conversation context with the power of pretrained large language models (LLMs). Through \textit{zero-shot, few-shot}, and fine-tuning experiments on four public conversation datasets, we show that for replies to the initial turn of a dialog, an LLM with 64B parameters is able to accurately expand over 70$\%$ of phrases with abbreviation length up to 10, leading to an effective keystroke saving rate of up to 77$\%$ on these expansions.
Including a small amount of context in the form of a single conversation turn more than doubles abbreviation expansion accuracies compared to having no context, an effect that is more pronounced for longer phrases. Additionally, the robustness of the models against typo noise can be enhanced through fine-tuning on noisy data.
\end{abstract}

\section{Introduction}

The prevalent paradigm of text entry on computing devices is sequential typing of characters. Word completion and prediction can theoretically save up to 40-50\% keystrokes when 3-5 predictions are provided \cite{trnka2008evaluating, fowler2015effects}. This reduces the motor and cognitive demand of entering text, especially on devices where typing is difficult, e.g., phones. 
In AAC use cases such as eye-gaze keyboards for severely motor-impaired individuals, the cost per keystroke is so high that there is a desire to save as many keystrokes as possible. Gaze-typing requires the user to precisely control the direction and timing of gaze for each keystroke, resulting in an extremely low text-entry speed of 8-10 words per minute and severely limiting real-time communication \cite{waller2019telling}. A text-entry paradigm with substantially higher keystroke saving rate (KSR) can reduce motor demand and thereby benefit AAC usage in real-time communication.

\begin{figure}
    \centering
    \includegraphics[width=0.9\columnwidth]{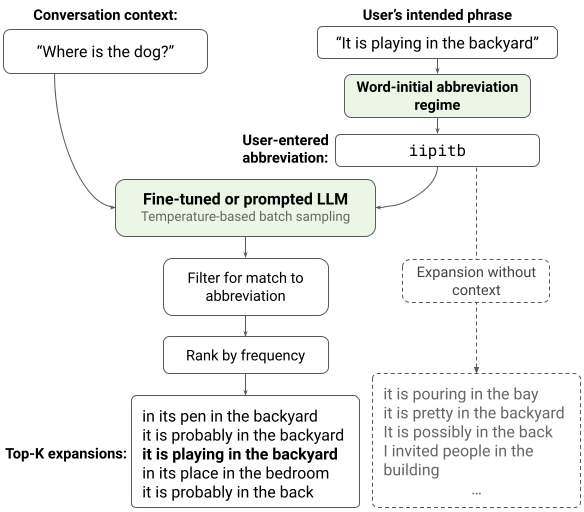}
    \caption{Our approach to abbreviation expansion based on an LLM with context compared to one without. The conversation context (e.g., a previous turn of conversation) along with the abbreviation of the intended phrase form the LLM's input.
    Sampled continuations from the model are filtered to discard those that do not match the abbreviation. Top-5 options after sorting by frequency are presented.
    }
    \label{fig:fig1}
\end{figure}

One potential paradigm is "SMS language", a spontaneously-evolved system for saving keystrokes in which each word is abbreviated as a single letter, such as in the well-known abbreviations {\it sg} for {\it sounds good} and {\it ttyl} for {\it talk to you later} \cite{anjaneyulu2013glossary}. SMS language features a high KSR (75-80\%), but is limited by its small closed set of common phrases of mostly six words or shorter. Its abbreviation scheme is not applied to longer or less frequent phrases because such abbreviations would be hard for the recipient to decipher. For example, the abbreviation {\it iipitb} is highly ambiguous and may represent many possible phrases, e.g., \textit{it is pouring in the bay} and \textit{it is pretty in the backyard} (see Figure~\ref{fig:fig1} for more examples). Some existing AAC systems support abbreviation expansion (e.g., \citet{TobiiAbbreviationExpansion}), but are limited by hardcoded, closed phrase sets.

The current study is based on the insight that although decoding open-set phrases from abbreviations is hard {\it without context} due to ambiguity, providing conversational context significantly constrains the space of likely phrases as shown by the example in Fig.\ref{fig:fig1} (\textit{it is playing in the backyard}). Hence we propose a high-KSR abbreviation scheme that focuses on conversational scenarios. We apply this scheme to three existing dialog datasets and create datasets for abbreviation expansion (AE).

This allows us to study whether LLMs, trained on web text including conversational data, can enable AE and benefit from added context. We take a 64B parameter LLM and compare zero-shot, few-shot, and fine-tuning performance on the AE task. Additionally, we simulate typing noise to study tolerance of the approach to typos. The main contributions of our work are:

\begin{enumerate}[wide]\itemsep-0.05cm
\vspace{-0.4cm}
    \item Demonstrating the potential of abbreviation expansion using LLMs aided by conversational context for highly-abbreviated text entry, while measuring the effects of different amounts of context and different dialog turns.
    \item Describing a high-KSR abbreviation scheme, a method for simulating typing noise, and conversation datasets based on these.
    \item Comparing \textit{zero-shot}, \textit{few-shot}, and model fine-tuning approaches for the AE task and their tolerance to typo noise.
\end{enumerate}

\section{Related Work}

\paragraph{Abbreviation expansion for text entry.}

Previous research on aiding text entry through AE used abbreviation schemes such as using only content words \cite{demasco1992generating}, discarding certain vowels and consonants \cite{shieber2007abbreviated}, and flexible letter saving schemes \cite{pini2010text,adhikary2021accelerating,gorman2021structured}. Spontaneous abbreviations schemes primarily omit vowels, repeating consonants, last characters, and spaces, and lead to modest KSR (e.g., 25-40\% in \citealt{willis2005probabilistic}, and 21\% in \citealt{adhikary2021accelerating}.) 
The low KSR of such schemes can be attributed  to the implicit need for a human reader to decode the phrases without significant cognitive burden.
N-gram models and neural language models (LMs) have been applied to expanding abbreviations for these relatively low-KSR schemes. By using LSTM models and context, ~\citet{gorman2021structured} achieve a word error rate of 1.5\%. \citet{adhikary2021accelerating} report a 24.2\% top-5 sentence error rate decoding abbreviations using an RNN to augment an n-gram LM.  
Our presented approach is a step towards using automation and context to expand abbreviations at a higher KSR that is close to that of SMS language. 

\paragraph{Large language model prompting and fine-tuning.}

Our approach builds on prior work on LLMs including few-shot prompting, fine-tuning, and conversation models \cite{raffel2019exploring,brown2020language,adiwardana2020towards,roller2020recipes}.  
We focus primarily on few-shot prompting \cite{brown2020language} and fine-tuning~\cite{ruder2021lmfine-tuning}. \textit{Few-shot} prompting uses a text description of a task along with a small number of examples for the task in the input text in order to elicit desired task responses from an LLM. In the \textit{zero-shot} scenario, no examples are provided. Prompting involves no updates to the model parameters. 
Model fine-tuning requires more data compared to prompting, but often leads to higher task accuracy than prompt engineering (e.g., \citealt{austin2021program, lester2021power}). For our AE task, data for fine-tuning 
can be synthesized from existing conversation datasets based on an abbreviation scheme (Sec.~\ref{sec:abbrev_regime}). Thus, we explore both  prompting and fine-tuning and compare their performance.

\paragraph{Assisting text entry with context.}

Textual contexts have been exploited to aid email writing ~\cite{kannan2016smart, chen2019gmail}. 
For text entry in AAC, \citet{wisenburn2008aac} demonstrated that providing noun phrases from a conversation partner's speech as selection options increases text-entry speed by 36.7\%. \citet{adhikary2019investigating} concluded that with currently-attainable accuracy of ASR, partner speech can be valuable in improving language modeling for AAC text entry.
\citet{shen2022kwickchat} used a fine-tuned GPT-2 model \cite{radford2019language} to expand bags of keywords into full phrases in conversational contexts based on the ConvAI2 dataset \citep{dinan2020second} and reported a KSR of 77\% at a word error rate threshold of 0.65. Our current study differs from the previous studies in the following aspects. First, we provide an abbreviation scheme to allow greater user control over the exact phrase structure and wording. Second, we performed detailed quantitative analysis of the combined predictive power of state-of-the-art LLMs and context awareness.

\section{Methodology}
\begin{table*}[!tb]
\centering
\small
\resizebox{\textwidth}{!}{
\begin{tabular}{l|ccc|ccc|ccc}
\toprule
\multicolumn{1}{c|}{{}} & \multicolumn{3}{c|}{{train}} & \multicolumn{3}{c|}{{dev}} & \multicolumn{3}{c}{{test}}  \\\cline{2-10}
\multicolumn{1}{c|}{{Dataset}} & \#conv. & \#examples &  Avg. tokens & \#conv. & \#examples & Avg. tokens & \#conv. & \#examples & Avg. tokens \\
\midrule
Turk Dialogues Corrected (TDC) & 859 & 5,154 & 54.4 $\pm$ 24.0 & 280 & 1,680 & 54.5 $\pm$ 24.3 & 280 & 1,680 & 55.0 $\pm$ 24.7 \\
Turk AAC (TAC) & 5,019 & 5,019 & 20.5 $\pm$ 4.3 & 559 & 559 & 20.9 $\pm$ 4.4 & 565 & 565 & 20.1 $\pm$ 4.0 \\
DailyDialog Corrected (DDC) & 11,188 & 87,170 & 101.1 $\pm$ 77.0 & 823 & 6,498 & 98.9 $\pm$ 72.3 & 772 & 5,852 & 96.7 $\pm$ 69.2 \\
Cornell Movie Dialog (CMD) & 66,848 & 244,798 & 68.3  $\pm$ 71.8 & 8,645 & 31,272 & 65.5 $\pm$ 67.4 & 7,444 & 27,429 & 69.8 $\pm$ 76.2 \\
\bottomrule
\end{tabular}
}
\caption{Summary of datasets with number of conversations (conv.), examples, and average tokens (mean $\pm$ 1 SD in number of SentencePiece tokens) used in our experiments for the context-aware AE task.
\label{tab:dataset_table}}

\end{table*}

\begin{table*}
\begin{center}
\small
\begin{tabularx}{\textwidth}{
|*{3}{Y|}} 
  \hline
  \textbf{Original dialog} & \textbf{AE example} & \textbf{AE example (noise $\sigma$=0.3)} \\ 
 \toprule
 \hline
\vspace{-0.42cm}
\messenger l {\footnotesize Would you like to sit down?}
\vspace{-0.28cm}
\messenger r {\footnotesize No, I'm fine standing up}
\vspace{-0.28cm}
\messenger l {\footnotesize Are you sure you don't
\\
want to sit down?} 
\vspace{-0.28cm}
\messenger r {\footnotesize Been sitting all day. Work was

just one meeting after another.}
\vspace{-0.28cm}
\messenger l {\footnotesize Oh, I'm sorry. I don't enjoy 

work days like that.}
\vspace{-0.28cm}
\messenger r {
\footnotesize It feels good to stretch 

my legs a bit.}
& 

\textbf{0-turn context:} Shorthand: \{wyltsd\}. Full: \{Would you like to sit down?\} 
\newline
\textbf{1-turn context:} Context: \{Would you like to sit down?\}. Shorthand: \{n,imfsu\}. Full: \{No, I'm fine standing up\} 
\newline
$\cdots$
\newline
\textbf{5-turn context:} Context: \{Would you like to sit down?\} \{No, I'm fine standing up\} \{Are you sure you don't want to sit down?\} \{Been sitting all day. Work was just one meeting after another.\} \{Oh, I'm sorry. I don't enjoy work days like that.\}. Shorthand: \{ifgtsmlab\}. Full: \{It feels good to stretch my legs a bit.\} 

&

\textbf{0-turn context:} Shorthand: \{wy\textcolor{red}{\ul{!}}tsd\}. Full: \{Would you like to sit down?\}
\newline
\textbf{1-turn context:} Context: \{Would you like to sit down?\}. Shorthand: \{n,i\textcolor{red}{\ul{n}}fsu\}. Full: \{No, I'm fine standing up\}
\newline
$\cdots$
\newline
\textbf{5-turn context:} Context: \{Would you like to sit down?\} \{No, I'm fine standing up\} \{Are you sure you don't want to sit down?\} \{Been sitting all day. Work was just one meeting after another.\} \{Oh, I'm sorry. I don't enjoy work days like that.\}. Shorthand: \{ifgtsm\textcolor{red}{\ul{o}}ab\}. Full: \{It feels good to stretch my legs a bit.\} \\ 
  \hline
\end{tabularx}
\end{center}
\caption{An example dialog and the generated AE examples without and with typo noise. The six-turn dialog is an excerpt from the train split of the TDC dataset. In the 3rd column, the typos in abbreviation are marked in \textcolor{red}{\ul{red}}.
\label{tab:ae_data_ex}}
\end{table*}

\paragraph{Abbreviation Scheme.}
\label{sec:abbrev_regime}


Our abbreviation scheme differs from previous studies in that we optimize for KSR and do not expect a human reader to be able to easily decode the abbreviations. Additionally, it offers the benefit that each given phrase is mapped to a fixed abbreviation. The detailed rules for abbreviating phrases are:
\vspace{-0.2cm}
\begin{enumerate}[wide, labelindent=5pt]\itemsep0em
   \item Each word is abbreviated as its initial letter, unless the word contains an apostrophe (i.e., contraction), in which case the word is split at the apostrophe and the initial letters from the splits are taken (e.g., {\it can't} -->  {\it ct}). This prevents abbreviations that are otherwise identical but semantically opposite (e.g., {\it can} vs. {\it can't}).  
   \item All letters in the abbreviation are lowercase.
   \item Arabic numerals in a sentence are preserved (e.g., {\it see you at 10 o'clock} --> {\it sya10oc}).
   
   \item Sentence-final punctuation are removed. Mid-sentence punctuation and special characters (e.g., {\it \#} and {\it \$}) are preserved to help constrain the structure of the sentence (e.g., {\it OK, but be quick.} -->  {\it o,bbq}).
\end{enumerate}


\subsection{Datasets for context-aware AE}
\label{sec:dataset}
We study modified versions of existing dialog datasets, which we converted for the context-aware AE task. We also describe how we simulate typos.


\paragraph{Datasets.}
Table~\ref{tab:dataset_table} summarizes the four datasets. We use their original train/dev/test splits in our experiments. 
The Turk Dialogues dataset \cite{vertanen2017towards} consists of crowd-sourced dialogs, each of which is exactly six turns in length. 
The dataset has typos and grammatical errors. We manually correct these and refer to the corrected dataset as \textbf{Turk Dialogues Corrected (TDC)}.\footnote{The corrected version is available in the file {\scriptsize\texttt{turk\_dialogues\_corrected.txt}} in Supplemental Data} We use three more datasets, \textbf{DailyDialog} \cite{li2017dailydialog}, a dataset of everyday conversations; the \textbf{Cornell Movie Dialogues (CMD)} \cite{danescu2011chameleons} based on movie scripts, and the \textbf{Turk AAC dataset (TAC)} \cite{vertanen2011imagination}. For evaluation on out-of-domain dialogs, we use the \textbf{TaskMaster-1 Self Dialogs (TMSD)} dataset \cite{byrne2019taskmaster}, a corpus of dialogs written by crowdworkers for task-oriented scenarios such as ordering pizza. TMSD is used only for evaluation and not for training or validation of the models. For DailyDialog, we remove 228 dialogues from the test split that are duplicate with conversations in the train split (see Appendix~\ref{sec:ddc_corrections}), which leads to what we call the \textbf{DailyDialog Corrected (DDC)} dataset. No correction is applied to the other datasets. 
The TAC dataset contains only isolated phrases without any conversational-turn context. Hence we use it only for training. In all of our experiments, we combine data from the training splits of all four datasets when fine-tuning models.
We perform evaluations on the TDC, DDC, CMD, and TMSD datasets. The TDC dataset is chosen as our primary 
benchmarks because of its strict six-turn dialog structure.


\textbf{Modifications for the AE task.} The above-mentioned datasets are typically used to study dialog generation. For our scenario, we convert each turn of the conversation in these datasets into the following canonical format:
\vspace{0.2cm}
\begin{tabular}{ r l } 
\toprule
\multirow{1}{4em}{Context:}   & \{Content of the contextual turn\}\\
\multirow{1}{4em}{Shorthand:} & \{Abbreviation of \textit{next turn}\}\\
\multirow{1}{4em}{Full:}      & \{Expanded content of \textit{next turn}\}\\
\hline
 \multirow{1}{4em}{Context:}   &  \{Would you like to sit down?\}\\
 \multirow{1}{4em}{Shorthand:} &  \{n,imfsu\}\\
 \multirow{1}{4em}{Full:}      &  \{No, I'm fine standing up\}\\
\bottomrule
\end{tabular}
%
%
For the AE task, the context consists of one or more previous dialog turns. When context is absent (e.g., for the opening turn), the context part is omitted. 
For a multi-turn dialog, the n$^{th}$ (1-based) example contains the first (n - 1) dialog turns as the context as well as the shorthand and the full form of the n$^{th}$ turn. Thus, a 6-turn conversation yields six examples for the AE task. When multiple sentences are present in a single turn, we use only the first sentence for expansion; when a turn is used as context, all available sentences are used.
Table \ref{tab:ae_data_ex} shows examples generated from all six turns of a dialog from TDC. 
Each dialog in the TDC, DDC, and CMD datasets yields several examples covering different amount of context.
We create only 0-context-turn examples for the TAC dataset since it contains only isolated phrases.

\paragraph{Text-entry noise in AE datasets.}
\label{sec:approach_noise}
As with our AE scheme, the introduction of noise to the datasets is also motivated by the AAC text entry use case, and in particular eye-gaze typing, which is error prone \cite{feit2017toward}. Here, misclicks occur frequently and must be taken into account when designing a gaze-driven text entry system. In order to simulate the noise, we model eye-gaze typing as uncorrelated 2D Gaussian distributions around the intended key \cite{azenkot2012touch}. 
\begin{figure}
    \centering
    \vspace{-0.4cm}
    \includegraphics[width=\columnwidth]{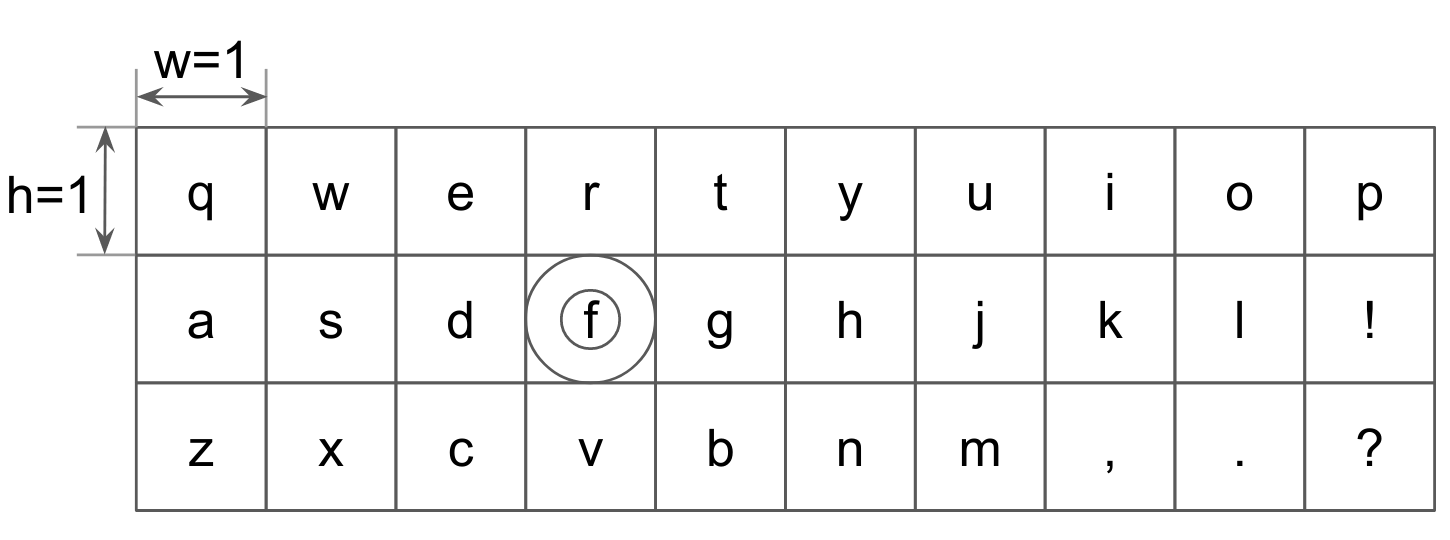}
    \caption{Keyboard layout for simulating noise in AE keypresses. The circles on the {\it f} key show $1\sigma$ around the mean for $\sigma \in \{0.3, 0.5\}$ in the 2D Gaussian distributions used to model typing noise.
    \label{fig:keyboard_layout_noise}}
    \vspace{-0.4cm}
\end{figure}
To simulate noise in the abbreviation input, we use a simplified rectangular-grid qwerty keyboard layout with 30 keys arranged in three rows and 10 columns. The keys are \begin{math}1\times1\end{math} squares with no gaps in between. The keystrokes for an intended key are drawn from 2D Gaussian distribution centered on the center of the intended key and standard deviations denoted $\sigma$ equal in the two spatial dimensions. To model different levels of noise, we use three values of $\sigma$: 0.0 (i.e., no-typo baseline), 0.3, and 0.5, which corresponds to 0\%, 13\%, and 44\% character error rates, respectively. Examples with simulated typos are shown in Table \ref{tab:ae_data_ex}.






\subsection{Large Language Model}
\label{sec:app_model}
One of our goals is to test whether \textit{zero-shot} and \textit{few-shot} prompting of LLMs are effective at the AE task without the need for supervised fine-tuning. Prompting is the method of eliciting desired task-specific responses from an LLM by including a natural-language description of the task and/or input-output examples of the task in the input string for an LLM, without altering the model's weights \cite{brown2020language}. Zero- and few-shot prompting differ in whether any examples are included in the prompt to the LLM. For this, we use a decoder-only Transformer language model~\cite{vaswani2017attention} from the LaMDA~\cite{thoppilan2022lamda} family of models. Our experiments are based on the 64B parameter model, unless otherwise specified. This model has $32$ Transformer layers, with $d_{model}=8192$, $d_{ff}=65536$, $h=128$, $dk = dv = 128$.
The model was pre-trained on 2.97B public web documents, Wikipedia, and dialogs.
The training data was tokenized with the SentencePiece vocabulary ~\cite{kudo2018sentencepiece} of size 32K.
We call this the \textbf{BaseLLM}.

We also developed \textbf{fine-tuned versions} of this model for the AE task. 
The fine-tuning uses examples in the format as shown in Table \ref{tab:ae_data_ex}. Since the BaseLLM is a decoder only model, and we use both the context and abbreviation as triggers to the model during inference, 
we modify the loss to only be calculated on the tokens of the AE target, i.e. the full form to be predicted in the pair of curly brackets after "Full:". For both training and inference, we split the characters in the abbreviation with spaces to force SentencePiece to use per-character IDs.  
We tune \footnote{Appendix~\ref{sec:appx_finetune} and~\ref{sec:appx_char_split} provide details on fine-tuning and discuss the effect of character splitting.} two models, \textbf{FT-LLM} on the combined AE datasets without typos, and \textbf{FTnoise-LLM} on the version with simulated typos. Both use early stopping on a dev set consisting of combined examples from the dev splits of TAC and TDC (Table \ref{tab:dataset_table}).

\section{Experiments}
\label{sec:experiments}

\paragraph{Models.} We use and compare the following models in our different experiment settings.

\textbf{Look-Up Table (LUT)}. As a straight-forward, non-ML baseline, we compile a dictionary of 375,298 sentence-level abbreviations from the train splits of the datasets in Table \ref{tab:dataset_table}. Each abbreviation maps to one or more phrases with their frequencies, leading to 447,249 unique abbreviation-sentence pairs. During evaluation, we map the query abbreviation to the top-5 expansion phrases (by frequency) by using the dictionary and breaking ties randomly.

\textbf{BaseLLM} (from Sec.~\ref{sec:app_model}).
We study the BaseLLM in the \textit{zero-shot} and \textit{few-shot} (specifically \textit{4-shot}) settings\footnote{The prompts are prefixed with the natural language instruction  ``Given acronym, write the full phrase.'' when there's no context or ``Given previous turn(s) of conversation and acronym of reply, write the full phrase.'' when there is context.
}.
The four examples are selected from the train split of the TDC dataset (see Appendix~\ref{sec:appx_4shot_ex}). 
We quantify the variability of the model on a sets of 856 4-example sequences from the train split of the TDC dataset. The best performing one on the dev set is denoted \textbf{BaseLLM$^{*}$}.

\textbf{FTnoise-LLM} tuned on simulated typos with noise level $\sigma = 0.3$ (see Appendix~\ref{sec:appx_noise.3_motiv}), and \textbf{FT-LLM} tuned on AE data without noise as described in Sec.~\ref{sec:app_model} are additional models we compare to.

\textbf{T5 encoder-decoder}. For comparison with smaller models, we use the T5 encoder-decoder \textbf{small} (60M), \textbf{large} (770M), and \textbf{3B} parameter models fine-tuned on AE data without noise, identical to FT-LLM.

We evaluate the fine-tuned models in the setting without any explicit natural language instructions (denoted ``no instr.'') unless mentioned otherwise. \noindent For all models, we perform random sampling with \textit{temperature=1.0} over the \textit{top\_k=40} candidates with the highest logits at each step. We decode \textit{128 samples} for each abbreviation unless otherwise specified. For each model and evaluation setting we report the standard deviations (SDs) of metrics over 3 repeated runs.

\paragraph{Studies.} For the BaseLLM, we study the variance in performance based on the prompt selection. For all the models, we sample multiple responses for each query, hence we study the effect of number of responses sampled on AE accuracy and latency. We also compare the performance of the models with varying amounts of conversation context and with no context. To study the effect of typos, we compare the performance of the models on the noise induced AE dataset. To measure the impact of model size on accuracy and latency, we also fine-tune and evaluate performance of the decoder-only LaMDA models with fewer than 64B parameters, specifically 4B, 8B, and 27B parameters. All these models were trained on the same data, so that the model size consitutes the only difference.
\begin{table}
\small
\begin{center}
\resizebox{\columnwidth}{!}{
\begin{tabular}{ c | c | c | c | c | c}
  \hline
  Abbv. & & & \\  
  length & \textbf{TDC (dev)} & \textbf{TDC (test)} & \textbf{DDC (test)} & \textbf{CMD (test)} & \textbf{TMSD (test)} \\ 
  \hline
 \hline
  1-2 & 85 (5.1\%) & 105 (6.2\%) & 166 (21.5\%) & 2,003 (26.9\%) & 176 (22.9\%) \\
  3-4 & 324 (19.3\%) & 293 (17.4\%) & 168 (21.8\%) & 1,753 (23.6\%) & 109 (14.2\%) \\
  5-6 & 454 (27.0\%) & 439 (26.1\%) & 152 (19.7\%) & 1,396 (18.8\%) & 113 (14.7\%) \\
  7-8 & 339 (20.2\%) & 376 (22.4\%) & 118 (15.3\%) & 851 (11.4\%) & 129 (16.8\%) \\
  9-10 & 221 (13.2\%) & 218 (13.0\%) & 64 (8.3\%) & 528 (7.1\%) & 111 (14.4\%) \\
  \hline \hline
  
  1-10 & 1,423 (84.7\%) & 1,431 (85.2 \%) & 668 (86.5\%) & 6,531 (87.8\%) & 638 (82.9\%) \\
  \hline
\end{tabular}
}
\end{center}
\caption{Datasets used for evaluation sliced by abbreviation lengths. Number of dialog turns in each range and their percentage (in parentheses) as compared to the total are noted.
\label{tab:eval_datasets}}
\vspace{-0.6cm}
\end{table}

\begin{table*}[h]
\centering
\resizebox{\textwidth}{!}{
\begin{tabular}{l|cc|cc|cc|cc|cc}
\toprule
\multicolumn{1}{c|}{{}}  &  
\multicolumn{2}{c|}{{TDC-test}} & 
\multicolumn{2}{c|}{{TDC-test$+$noise ($\sigma$=0.3)}} &
\multicolumn{2}{c|}{{DDC-test}} &
\multicolumn{2}{c|}{{CMD-test}} &
\multicolumn{2}{c}{{TMSD-test}} \\\cline{2-11}
\multicolumn{1}{c|}{{Model}} & Acc.@5 & BLEU@5 & Acc.@5 & BLEU@5 
& Acc.@5 & BLEU@5
& Acc.@5 & BLEU@5
& Acc.@5 & BLEU@5 \\
\midrule
Look-Up Table (LUT)  & 14.3 $\pm$ 0.2 & 23.6 $\pm$ 0.1 & 10.5 $\pm$ 0.0 & 15.8 $\pm$ 0.7 & 48.1 $\pm$ 0.2 & 55.4 $\pm$ 0.3 & 30.9 $\pm$ 0.1 & 39.2 $\pm$ 0.1 &  29.3 $\pm$ 0.1 & 34.7 $\pm$ 0.1\\
\hline
T5-small (60M) & 42.7 $\pm$ 0.5 & 59.9 $\pm$ 0.1 & 21.2 $\pm$ 0.1 & 36.1 $\pm$ 0.3 & 69.1 $\pm$ 0.5 & 78.1 $\pm$ 0.6 & 38.7 $\pm$ 0.0 & 50.4 $\pm$ 0.1 & 50.7 $\pm$ 0.3 & 64.8 $\pm$ 0.5 \\
T5-large (770M)  & 55.2 $\pm$ 0.6 & 68.6 $\pm$ 0.4 & 27.3 $\pm$ 0.6 & 40.9 $\pm$ 0.3 & 74.2 $\pm$ 0.1 & 81.7 $\pm$ 0.1 & 41.2 $\pm$ 0.0 & 52.6 $\pm$ 0.1 & 57.1 $\pm$ 0.1 & 70.1 $\pm$ 0.2 \\
T5-3B (3B)  & 59.4 $\pm$ 0.4 & 72.8 $\pm$ 0.1 & 26.9 $\pm$ 0.8 & 41.9 $\pm$ 0.7 & \textbf{77.6 $\pm$ 0.5} & \textbf{83.9 $\pm$ 0.5} & 43.5 $\pm$ 0.1 & 54.8 $\pm$ 0.2 & 59.5 $\pm$ 0.2 & 72.5 $\pm$ 0.3 \\
\hline
BaseLLM$^*$ 64B (best, $4shot$) & 43.7 $\pm$ 1.2 & 54.9 $\pm$ 0.5 & 38.1 $\pm$ 0.1 & 42.0 $\pm$ 0.5 & 38.4 $\pm$ 0.4 & 43.3 $\pm$ 0.6 & 22.5 $\pm$ 0.2 & 25.9 $\pm$ 0.1  & 32.0 $\pm$ 0.7 & 36.2 $\pm$ 0.3\\
FT-LLM 64B (no instr.) & \textbf{74.4 $\pm$ 1.0} & \textbf{81.8 $\pm$ 0.8} & 44.5 $\pm$ 0.7 & 55.0 $\pm$ 0.3 & 75.1 $\pm$ 0.6 & 82.1 $\pm$ 0.6 & \textbf{48.1 $\pm$ 0.1} & \textbf{57.9 $\pm$ 0.2}  & 62.0 $\pm$ 0.3 & 73.9 $\pm$ 0.2\\
FTnoise-LLM 64B (no instr.)  & 72.3 $\pm$ 0.9 & 81.1 $\pm$ 0.5 & \textbf{60.9 $\pm$ 0.3} & \textbf{71.4 $\pm$ 0.5} & 74.8 $\pm$ 0.4 & 82.1 $\pm$ 0.3 & 47.5 $\pm$ 0.1 & 57.3 $\pm$ 0.1 & \textbf{63.3 $\pm$ 0.1} & \textbf{74.4 $\pm$ 0.2}\\
\bottomrule
\end{tabular}
}
\caption{Comparing models (from Sec.~\ref{sec:experiments}) on the AE task on turn-2 given turn-1 as context. We report accuracy and BLEU score at top-5, as percentages, std. dev. computed on 3 runs. Higher is better, values in \textbf{bold} are highest in each column.
\label{tab:tab_res_table}}
\end{table*}

\paragraph{Evaluation.}
We only evaluate on conversation queries with abbreviation length $\le10$ characters.
This encompasses the majority (85$\%$) of the dialog turns from the original dataset (Table \ref{tab:eval_datasets}). Where applicable, we prepend the following natural-language instruction to the model input for the AE task: \textit{"Given previous turn(s) of conversation and acronym of reply, write the full phrase."}

Before calculating performance metrics, we filter the model's responses: we remove sentence-final punctuation, standardize whitespace to one space, lower-case, de-duplicate, and filter for precise match of the abbreviation. The responses that pass the filtering are sorted by descending count. For evaluation with noise, we do filtering to allow matches to nearby characters on the keyboard. 

\paragraph{Metrics.}\textbf{Accuracy} measures whether any response expansion exactly matches the ground truth (with standardized letter-casing and whitespace, and discarded final punctuation). Additionally, we measure  \textbf{BLEU} score~ \cite{papineni2002bleu} using the SacreBLEU library \cite{post2018call} as a more fine-grained metric for the similarity between AE options and the ground truth. For both metrics, we report performance in the top-5 responses after they are sorted based on frequency.

\textbf{Key Stroke Savings (KSR)} measures the number of saved keystrokes compared to the full length of the phrase. Note, however, that AE succeeds only for a subset of the cases, while for others 
the top-5 options do not contain the intended phrase. Hence we compute two types of KSR: 

\textbf{KSR$_{all}$}, computed on all phrases, is defined as 
    \begin{equation}
    \label{eq:ksr}
    \scriptsize
      \scriptsize{KSR_{all}}=\begin{cases}
        \left(1 - \frac{L_{abbrev}}{L_{full}}\right) \times 100, & \text{\scriptsize if in top-5}.\\
        \left(1 - \frac{L_{abbrev} + L_{full}}{L_{full}}\right) \times 100, & \text{\scriptsize otherwise}.\\
      \end{cases}
    \end{equation}
    where $L_{abbrev}$ and $L_{full}$ are the character lengths of the abbreviation and full phrase, respectively. In other words, if a phrase has a matching option in the top-5, we calculate the KSR as the percentage of keypresses saved by using the abbreviation. If the ground truth is not in top-5, we add a penalty term ($L_{full}$) to account for the need to enter the phrase by starting anew character-by-character, leading to a \textit{negative} KSR. $KSR_{all}$ is calculated by averaging over all phrases in an experiment.
\textbf{KSR$_{success}$}, is calculated by averaging over only the subset of phrases with exact matches and uses the first case in Equation \ref{eq:ksr}.

\section{Results}
We present the main results comparing the models on all datasets in Table~\ref{tab:tab_res_table} and then highlight results from specific experiments.

\paragraph{The accuracy of LLMs at expanding word-initial abbreviations is enhanced by fine-tuning.}
Table~\ref{tab:tab_res_table} compares the performance of all the models on the abbreviation expansion (AE) task\footnote{Appendix Tab. \ref{tab:tab_tdc_dev_table} reports performance on dev split of the TDC (TDC-dev) which was used for hyperparameter tuning.}. 
The data shown in the table are for AE on the 2$^{nd}$ turn of a dialog that utilizes the 1$^{st}$ turn as the context, which focuses on our main hypothesis regarding the effect of context on AE.

It's noteworthy that the BaseLLM$^*$, which has seen just four examples in its prompt (unlike the other models), shows performance that exceeds the look-up table (LUT) baseline in many cases, demonstrating the versatility of LLMs. The higher scores of the LUT on DailyDialogs (DDC) and Cornell Movie Dialogues (CMD) datasets are indicative of the high percentage of similar phrases in the train and test sets of the datasets.
Unsurprisingly, the fine-tuned models (FT-LLM , FTnoise-LLM, and T5 models) far outperform even the best \textit{4-shot} BaseLLM$^*$, achieving 74-77\% top-5 exact-match accuracy on the TDC and DDC datasets in the absence of typo noises. The accuracies are lower on the CMD dataset (comprised of movie scripts.)  The out-of-domain evaluation on the TaskMaster Self Dialogs (TMSD) dataset also showed accuracies lower than the TDC and DDC datasets, but higher than the results from the CMD dataset. 

\paragraph{Fine-tuning and tolerance to noise.} For conditions that involve simulated typo noise in the abbreviation input, FTnoise-LLM shows superior performance compared to other models (see the column "TDC-test + noise" in Table \ref{tab:tab_res_table}.) Interestingly, the performance of the BaseLLM$^*$ doesn't drop as much as any of the fine-tuned models - T5 or FT-LLM - in this setting. However, while FT-LLM still outperforms BaseLLM on the noisy abbreviations, the smaller T5 models fail to do so.


\paragraph{Context is critical for AE accuracy}
\begin{figure}
    \centering
    \includegraphics[width=\columnwidth]{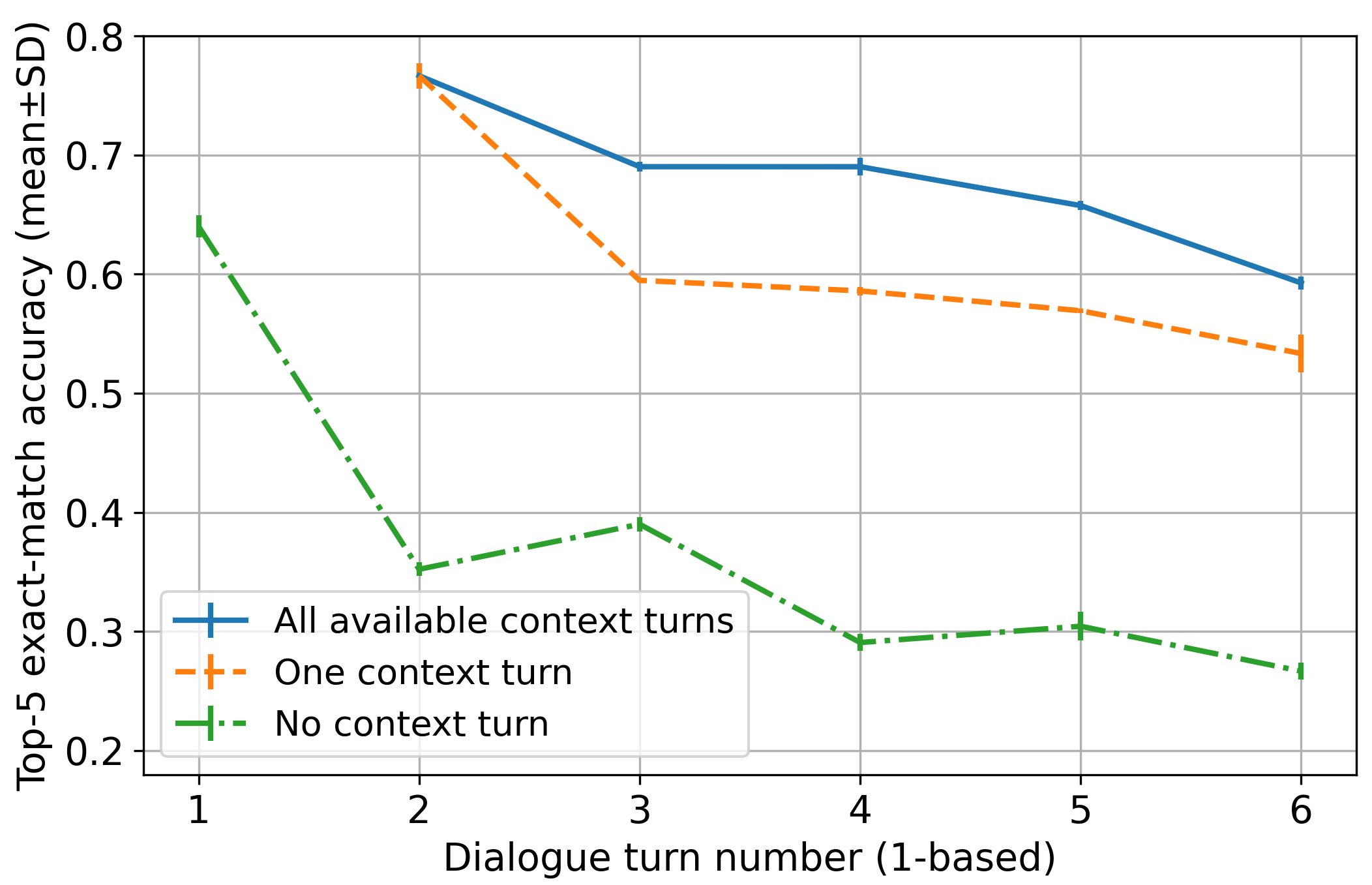}
    \caption{AE accuracy of FT-LLM, evaluated (inference only) with different amounts of input context (different curves) on different dialog turns (x-axis) on the TDC dev set. With all turns as context (solid blue curve) or just the previous turn as context (dashed orange curve), the model considerably outperforms the setting where no context is provided (dot-dash green) with the abbreviation query.
    }
    \label{fig:fig_finetuned_dev}
\end{figure}

Figure~\ref{fig:fig_finetuned_dev} show how the AE accuracy of FT-LLM varies when different amounts of context from previous turns of the conversation are provided. Compared to having no context (dash-dotted curve), 
including just one previous turn of context (dashed curve) approximately doubles accuracy. 
Using the full context (all dialog turns from the 1$^{st}$ to the $(n$-$1)^{th}$, solid curve) leads
to further improvements indicating that prior turns carry useful information for the AE task.


Compared to the 1$^{st}$ turn, AE under no context on subsequent turns (2$^{nd}$-6$^{th}$) shows significantly worse accuracy. This is due to the fact that the first turn consists of conversation starters that are easier to predict without context. Overall, irrespective of context, the accuracy of AE decreases as the number conversation turns increases, indicating increasing difficulty in predicting the full phrases from the abbreviation as the dialogs progress.
However, including full context during inference still achieves accurate expansions for 60$\%$-70$\%$  of the cases on the later turns.

\begin{figure}
    \centering
    \includegraphics[width=\columnwidth]{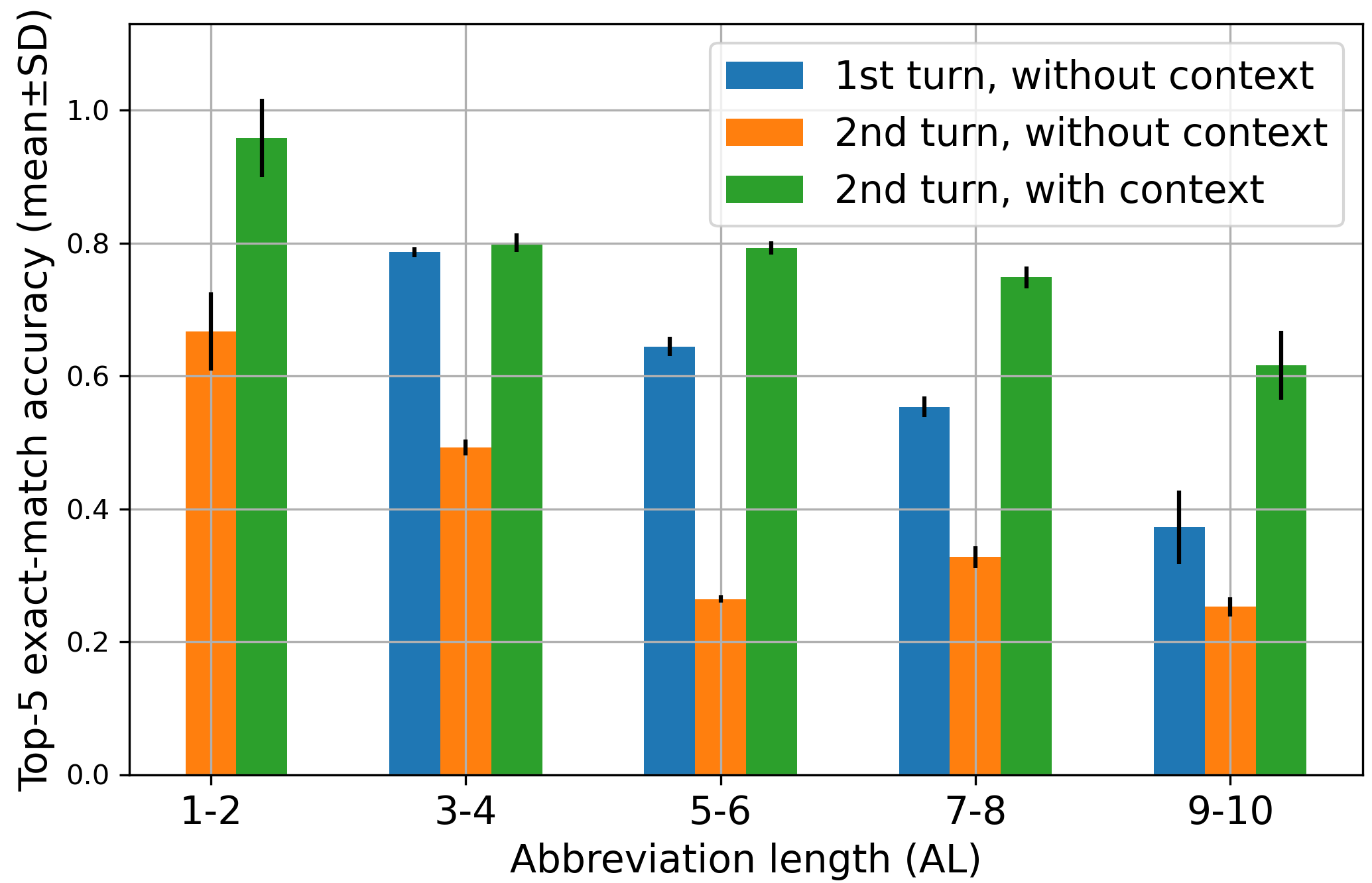}
    \caption{AE accuracy as a function of abbreviation length (AL). The results shown are from FT-LLM evaluated with no prompt. Different colors of bars show AE on the 1st and 2nd turns of the dialog in the TDC dev split, with 0 and 1 previous turn as the context. The 1-2 bin contains no 1st-turn examples.}
    \label{fig:fig_al_breakdown}
\end{figure}

\textbf{Effect of context is more pronounced on longer abbreviations.}
When performance is sliced by the abbreviation length (Figure~\ref{fig:fig_al_breakdown}), accuracy without context decreases sharply and nearly monotonically with increasing abbreviation length, regardless of whether it's the opening turn or the 2nd turn. With context however, the accuracy remains higher and decreases more slowly with abbreviation length, extending the approximately 80\% or higher accuracy into longer phrase lengths.

\paragraph{The variability and usefulness of few-shot prompts decreases after model tuning.}

\begin{table}[h]
\centering
\begin{center}
\begin{tabular}{||l l l||} 
 \hline
 Acc.@top-5 & BaseLLM & FT-LLM  \\ [0.5ex] 
 \hline\hline
 4-shot prompt & 31.71 $\pm$ 4.83 & 74.43 $\pm$ 1.79 \\
 0-shot prompt & 37.10 $\pm$ 1.38 & 77.10 $\pm$ 0.38 \\
 No instr. & 14.00 $\pm$ 1.01 & 76.65 $\pm$ 1.06 \\ [1ex] 
 \hline
\end{tabular}
\end{center}
\caption{Mean and standard deviation of Accuracy@top-5 for the BaseLLM and FT-LLM over 856 different 4-shot prompts from the TDC train set, 3 repeated runs under 0-shot prompts (instruction only) and No instr. (i.e., neither instructions nor examples), based on AE on turn-2 given turn-1 as context.
\label{tab:variance_prompt}}

\end{table}
Here we focus on how much the LLM benefits from prompting before and after fine-tuning. The first row of Table~\ref{tab:variance_prompt} compares AE accuracies from different 4-shot prompts on the TDC dataset for BaseLLM and FT-LLM. We use the 856 example abbreviation-expansion pairs from the train split of the TDC dataset, using four conversation examples for the prompt at a time. The BaseLLM shows a large variance in performance depending on the selected examples in the prompt by as much as $SD=4.83$. The best 4-shot prompt for BaseLLM outperforms the 0-shot prompt, despite the fact that the average 4-shot prompt accuracy is lower. Therefore for BaseLLM we report the results from the best 4-shot prompt (BaseLLM$^*$). By contrast, the fine-tuned model (FT-LLM) shows significantly lower prompt-related variance ($SD=1.79$) in addition to a 2.3-fold increase in the mean accuracy. Moreover, FT-LLM is able to perform the AE task with only a natural-language prompt without examples (0-shot prompt) and even without any instruction (``No instr.'') at average accuracies that are more than 1 SD above that of 4-shot prompting. The ``No instr.'' setting is attractive due to its simplicity (no need to search for or hand-engineer a prompt) and reduced latency (due to shorter input prefix lengths). 
Given these results, we use the ``No instr.'' as the default setting and for all other experiments on FT-LLM and FTnoise-LLM. 


\paragraph{Increasing number of decoded samples improves accuracy at the cost of latency.}
\begin{figure}
    \centering
    \includegraphics[width=\columnwidth]{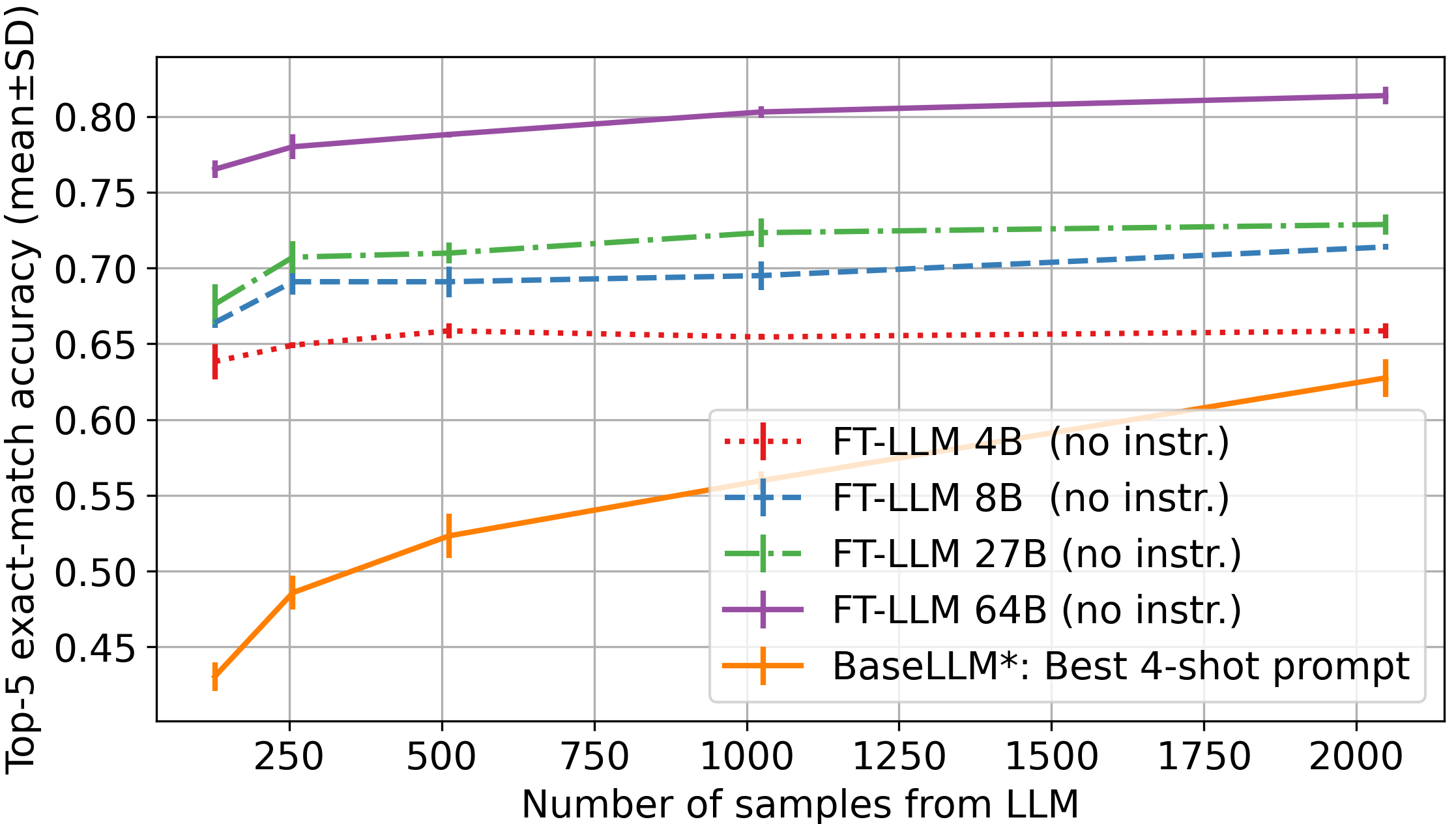}
    \caption{Increasing number of samples from the LLMs improves top-5 exact-match accuracy. FT-LLMs, even with fewest samples and smallest model size, outperform the BaseLLM$^*$.
    }
    \label{fig:num_samples_acc_lat}
\end{figure}
Latency is important for interactive text-entry applications. During sampled decoding, the LLMs generate 128 continuations of length 16 tokens for a batch of prefix length 256 with a median latency of 0.568 s (interquartile range: 0.16 s). 

This latency is close to typical dwell time of eye-gaze keyboards \cite{majaranta2007text} and hence could be acceptable for the eye-gaze typing use cases. Figure~\ref{fig:num_samples_acc_lat} shows the effect of increasing the number of continuations sampled from the LLMs. As expected, increasing sample count from 128 to 2048 improves top-5 accuracy for both BaseLLM* (with 4-shot prompts) and FT-LLM (no instr.). Improved accuracy comes at the cost of increased latency.\footnote{Note, that it is possible to cut down latency by parallelizing sampling, however this might increase hardware requirements at inference time.} BaseLLM benefits significantly more from increasing sample count than FT-LLM.

\paragraph{Comparison of model sizes}
Figure~\ref{fig:num_samples_acc_lat} also  compares fine-tuned models of different sizes (4B, 8B, 27B, and 64B). With model fine-tuning, the accuracy increases monotonically with increasing number of parameters. Interestingly, even with the fewest samples (128), fine-tuned models of all sizes outperform the larger (64B) model under \textit{few-shot} learning. Amongst the encoder-decoder T5 models (Table~\ref{tab:tab_res_table}) larger models significantly outperform smaller ones. As observed for the decoder-only models, the smaller fine-tuned T5 models outperform the few-shot BaseLLM in almost all cases except when the input consists of typos.

\paragraph{Keystroke saving rates.}

\begin{table}
\small
\begin{center}
\resizebox{\columnwidth}{!}{

\begin{tabular}{ c | c | c | c}
  \hline
  \multirow{2}{0.22\columnwidth}{Dataset-split} & & \\  
  & AE task & \textbf{$KSR_{all}$} & \textbf{$KSR_{success}$}\\ 
  \hline
 \hline
  TDC-test & 1st turn (no context) & 37.1 $\pm$ 0.19 & 76.8 $\pm$ 0.04 \\
           & 2st turn (with context) & 49.0 $\pm$ 0.99 & 73.5 $\pm$ 0.03 \\
  DDC-test & 1st turn (no context) & 20.0 $\pm$ 1.15 & 74.6 $\pm$ 0.04 \\
           & 2st turn (with context) & 49.0 $\pm$ 0.60 & 72.9 $\pm$ 0.04 \\
  \hline
\end{tabular}
}
\end{center}
\caption{KSR computed on all phrases and only phrases with matching AE options. The data in this table is computed on the results from FT-LLM.
\label{tab:ksr}}
\end{table}

KSR can be considered as a proxy measure of usability of the approach for AAC use-cases. 
$KSR_{success}$ values are in the range of 73-77\% for the 1st and 2nd turns of dialogs in the TDC and DDC datasets (Table~\ref{tab:ksr}), indicating that our proposed AE scheme does indeed lead to high KSRs. Values of $KSR_{all}$ are lower, reflecting the penalties for when a perfect match is not achieved. 
However, with context, $KSR_{all}$ approaches 50\% and is higher compared to no context (20\%-37\%).
Note that $KSR_{all}$ is extremely conservative as it does not consider (a) the possibility of using the information already contained in the abbreviation to "recover from AE failure" (e.g., by letting the user specify a word and invoke the LLM again) or (b) the fact that word completion and prediction may still be utilized even if the user falls back to sequential text entry.

\paragraph{Fine-tuning with noise improves typo tolerance.}

\begin{figure}
    \centering
    \includegraphics[width=\columnwidth]{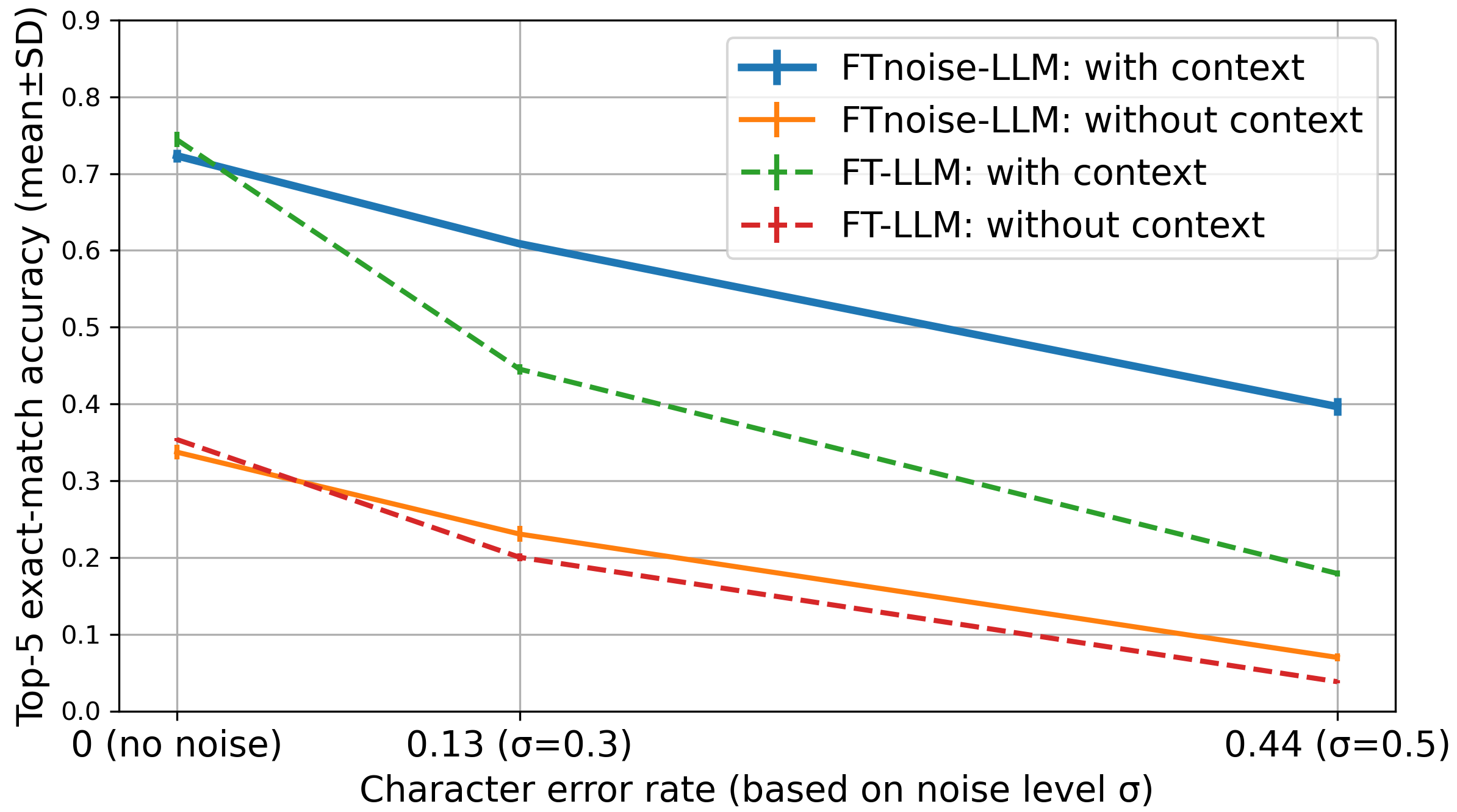}
    \caption{AE accuracy with and without typo noise in the input abbreviation. We compare the accuracies of the models fine-tuned without and with noise. Each curve shows the average top-5 accuracy in the 2nd turns of the dialogs in the test split of the TDC dataset. 
    }
    \label{fig:fig_typotolerance}
\end{figure}

Figure \ref{fig:fig_typotolerance} compares the AE accuracies of LLMs fine-tuned with and without noise (FTnoise-LLM and FT-LLM). 
While both models show decreasing AE accuracies with increasing amounts of typos, FTnoise-LLM is much more robust showing lesser drop in performance. Further, on noise-free inputs ($\sigma$=0), FTnoise-LLM shows only slight accuracy deterioration compared to FT-LLM.
We also find that typo tolerance, for both FT-LLM and FTnoise-LLM, is more pronounced with context than without.

\paragraph{Cross-domain generalization.}
We use the TMSD dataset to compare and evaluate the performance of models on conversation domains not seen in training. In Table~\ref{tab:tab_res_table} we can observe that few-shot prompting does fall behind the simple Look-Up Table baseline on DDC and CMD datasets. However, when we evaluate the models on cross-domain TMSD dataset of dialogs we can observe that the fine-tuned and few-shot models do generalize better to unseen domains and perform better than the baseline look-up.

\section{Discussion}

\paragraph{Qualitative analysis of AE failures.}

As indicated by the relatively high BLEU scores in Table~\ref{tab:tab_res_table} ($>80\%$), there are many expansions in the top-5 options that are "near misses". Appendix Table \ref{tab:ae_near_misses} shows a few examples of such near misses, in which the options differ from the the ground-truth phrase by only a semantically-similar word (e.g., ``yes'' vs. ``yeah'', ``head out'' vs. ``head over''.) Future studies need to investigate the frequency and UX acceptability of such near-miss AE options. But their existence implies that exact-match accuracy reported above slightly underestimates the practical effectiveness of the models. Another category of AE failures involve phrases that contain certain proper nouns. The last four examples in Table \ref{tab:ae_near_misses} show such cases in which the model correctly expands all the words but a proper noun. When such errors occur, the model tends to predict more common proper nouns, which is likely a reflection of the higher frequency of the predicted nouns in the model's pre-training and fine-tuning datasets.

\paragraph{The benefit of AE relative to sequential text entry.}

Word completion and prediction incur scanning cost: users scan the options in order to determine whether any of them match their intention, which has a detrimental effect on speed that needs to be overcome by the high quality of the options \cite{trnka2009user}. Although the speed of AE-based text entry remains to be quantified in future studies, we point out that: (1) AE removes overhead of scanning for options in between keystrokes, (2)  there are fewer characters to examine or correct when typing, both of which may offer speed-ups in addition to the higher KSR afforded by AE.

Although the current study is motivated by and focuses on the AAC use case, our paradigm of abbreviated text entry may be applicable to text input on touch screens as well. The AE approach of the current study can be regarded as a variation of contextual prediction of user text \cite{kannan2016smart, chen2019gmail} that affords greater flexibility in message content at the trade-off of requiring specification of the message with a small number of keystrokes. 

\paragraph{Future directions.} 
We found fine-tuning to be significantly better than prompting in terms of (a) accuracy (for both scenarios with and without typo-noise) and also (b) exhibit lower latency as we achieve better results with fewer samples. Future work should investigate the differences in latencies between the encoder-decoder architecture and decoder-only models. For training efficiency, instead of fine-tuning, it will also be worth investigating strategies such as prompt tuning~\cite{lester2021power} that continue to keep the model frozen, but learn some additional parameters for the task.

Even in the best case scenario models can fail to find accurate expansions\footnote{see Appendix~\ref{sec:appx_recovery_failure} for analysis} among the top-5 options. Recovering from such failures is important for AAC use cases. Future studies should consider options for partial specifications of one or more words or selection of some words from the available options. Once the recovery from failure is proven in offline analysis, user studies are required to validate and quantify the actual benefit of the AE text-entry paradigm in lab and real-life settings. Integration with UI approaches is also an essential direction, e.g., speeding up eye-gaze typing such as cascading dwell time and dwell-free paradigms \cite{mott2017improving, kristensson2012potential}.

\section{Conclusion}
In this work we proposed a high-KSR form of abbreviation expansion to dramatically save keystrokes for severely-disabled users. We use it to synthesize three datasets for the AE task. Based on extensive experiments using few-shot prompting and model tuning we demonstrate that across the datasets, fine-tuned LLMs can accurately predict expansions for 48-77\% of phrases that are replies to initial turns of dialogs and exhibit KSRs in the range of 73-77\% for the correctly predicted expansions, thus pointing at a promising direction for future user studies of contextual and abbreviated text entry based on LLMs.
Models evaluated with conversation context show significantly higher accuracy than without, thus supporting our hypothesis that context is the key to effective abbreviated text entry in conversational settings. Furthermore, fine-tuning with simulated typos substantially improves tolerance to noise in abbreviation.


\section{Acknowledgements}
We would like to thank Shumin Zhai and Michael Terry for feedback on a draft of this work, Yanping Huang for pointers on model inference, as well as James Stout, Bob MacDonald, Julie Cattiau, and Maarten Bosma for their support. We are grateful to Team Gleason for their active involvement and feedback in the development of this work. 

\section{Ethical Considerations, Limitations, and Societal Impact}

Accelerating augmentative and alternative communication (AAC) can enhance quality of life of people with extremely limited mobility by facilitating increased social participation and independence~\cite{calgari2013}. While the benefits of AE may be large for this population, we note that this approach may have risks. 

The primary risk of AE is errors in expansions that substantially misrepresent the intent of the speaker in a way that might cause harm to themselves or others (e.g., failure to correctly convey critical health information, insertion of offensive language.) The abbreviation expansions may also reflect biases in the underlying language model (e.g., perpetuating stereotypes by more frequently suggesting male pronouns than female, \citealt{weidinger2021ethical}.)

A more subtle risk is when expansions miss the ground-truth phrase closely (see Table~\ref{tab:ae_near_misses}), which may accurately convey content but reduce the speaker's sense of autonomy and authentic self-expression. Prior work (e.g., \citealt{kane2017cscw}) has shown that people with ALS highly value AAC that preserves and facilitates authentic identity expression. Providing speakers with multiple AE options to choose from and requiring user confirmation before voicing an expansion are design options that can mitigate these risks. Model fine-tuning to improve safety or personalization to the end-user's communication style are additional risk-mitigation approaches.  

Beyond enhancing communication speed, another intended benefit of AE is the potential to reduce fatigue associated with gaze-based AAC by reducing keystrokes; however, a risk of our system is that if errors in AE are frequent for a given user (perhaps due to eye tracker miscalibration or long-tail abbreviation use) then these savings could be outweighed by the need to correct errors, inadvertently increasing fatigue. User studies to better understand error rates in practice, as well as future work designing interfaces to simplify AE error correction, are important for minimizing this risk. Similarly, our abbreviations scheme's simple design based on first letters aims to minimize cognitive load; however, user studies with the target population using instruments such as NASA's Task Load Index\footnote{\url{https://humansystems.arc.nasa.gov/groups/tlx/}} would be required to verify that AE does not cognitively strain end-users.

\bibliography{references}
\bibliographystyle{acl_natbib}

\newpage
\appendix
\section*{Appendix}

\section{Removal of duplicate dialogs from the DailyDialog dataset}
\label{sec:ddc_corrections}

We observed that the DailyDialog dataset \cite{li2017dailydialog} contains a significant number of dialogs in its dev (validation) and test splits that are identical or nearly identical to the dialogs found in its train split. We determined two dialogs to be duplicate by using the following criterion:

\begin{enumerate}\itemsep0em
   \item If both dialogs consist of the same number of turns and the corresponding turns are all identical (case-insensitive), or
   \item If both dialogs consist of the same number of turns and there are three or more turns at which both dialogs contain identical text (case-insensitive).
\end{enumerate}

See the file daily\_dialog\_deduplications.csv in Supplemental Data for a list of the 177 dialogs in the dev split and the 228 dialogs in the test splits that are found to be duplicates with the train split and hence are removed from our DailyDialog Corrected (DDC) dataset.

\begin{table*}
\begin{center}
\small
\begin{tabularx}{\textwidth}{p{0.02\textwidth}p{0.23\textwidth}p{0.09\textwidth}p{0.30\textwidth}p{0.30\textwidth}}
  \hline
  \# & Context & Abbreviation & Ground truth & Non-matching expansion options \\ 
 \hline

1 & Awesome! My favorite weather! & \it{swhottwp} & Shall we head over to the water park? & \textbf{shall we head out to the water park} \\ 

\hline

2 & Can we go out for a drive? & \it{ygstc} & Yeah go start the car & \textbf{yes go start the car} \\ 
& & & & yes go straight to church \\ 
& & & & yes go settle the children \\ 
& & & & yeah get some tunes cranked \\ 
& & & & yes go straight to chicago \\

\hline

3 & i took a lot of courses, such as philosophy, logic, ethics, aesthetics, etc & \it{wcdylb} & which course did you like best & what courses do you like best \\ 
& & & & what courses did you like best \\ 
& & & & what course do you like best \\ 
& & & & what course did you like best \\ 
& & & & \textbf{which courses did you like best} \\

\hline

4 & it's hard to be optimistic about things with the way the economy's headed... the trade deficit is getting larger, consumption's down, i really think we're headed for a recession & \it{tehbsfawn} & the economy has been stagnant for a while now & the economy has been slowing for a while now \\
& & & & \textbf{the economy has been sluggish for a while now} \\
& & & & the economy has been strong for a while now \\
& & & & the economy has been slow for a while now \\
& & & & the economy has been suffering for a while now \\

\hline

5 & What is your name? & \it{mnir} & My name is Rey & my name is robert \\ 
& & & & my name is rebecca \\ 
& & & & my name is richard \\
& & & & my name is rose \\ 
& & & & my name is roy \\

\hline

6 & hey, isabelle... & \it{l} & Logan & lisa \\ 
& & & & linda \\ 
& & & & look \\
& & & & lillian \\ 
& & & & liz \\

\hline

7 & so, paula, where are you from & \it{imfc,o} & i'm from canada, originally & i'm from china, ok \\ 
& & & & i'm from california, originally \\ 
& & & & i'm from california, ok \\ 
& & & & i'm from california, okay \\ 
& & & & i'm from california, obviously \\

\hline 

8 & hey sandra, what's wrong? you look furious & \it{ivhiwt} & i've had it with Tim & i've had it with this \\
& & & & i've had it with them \\
& & & & i've heard it was true \\
& & & & i've had it with that \\
& & & & i've had it with these \\

\hline

\hline
\end{tabularx}
\end{center}
\caption{Examples of failed AE. Examples \#1-4 show AE options that miss the ground-truth phrase closely. The cases highlighted in boldface have near identical meaning to the ground truth, but differ only in details of a single word. Examples \#5-8 show AE options that match the ground truth except for the a proper noun.
\label{tab:ae_near_misses}}
\end{table*}

\section{4-$shot$ examples for BaseLLM$^*$}
\label{sec:appx_4shot_ex}
We select four consecutive dialogues from the 859 examples from train split of the TDC dataset \cite{vertanen2017towards} while varying the starting conversation, which yields $859 - 4 + 1 = 856$ different 4-shot prompt sets.

\section{Tuning on noisy data vs. accuracy}
\label{sec:appx_noise.3_motiv}
Preliminary experiments have shown that $\sigma=0.3$ is a good trade-off between accuracy gains on noisy data and losses on non-noisy data.

\section{Model fine-tuning details}
\label{sec:appx_finetune}
Our model fine-tuning uses the AdaFactor optimizer~\cite{shazeer2018adafactor}. The nominal batch size 16 is made more efficient through example packing~\cite{raffel2019exploring}, leading to an average effective batch size of approximately 200 examples under a maximum sequence length of 1024 tokens. We used TPUv3s \cite{jouppi2018motivation} with a configuration of 4x8 for the LLM fine-tuning. 
Our fine-tuning recipe applies a constant, low learning rate of $5\times10^{-5}$ and a dropout rate of $0.2$, which helps to prevent early overfitting. Early stopping is based on a dev set consisting of combined examples from the dev splits of the TAC and TDC datasets. We find the best checkpoint after 2100 and 1800 training steps for the FT-LLM and FTnoise-LLM models, respectively, which amounts to approximately 1-1.2 epochs of training. We ran a small set of hyperparameter tuning experiments, varying batch size, learning rate and dropout and chose the best setting based on the TAC$+$TDC dev set.

\begin{table}[h]
\centering
\resizebox{\columnwidth}{!}{
\begin{tabular}{l|cc}
\toprule
\multicolumn{1}{c|}{{}} & \multicolumn{2}{c}{{TDC-dev}} \\
\multicolumn{1}{c|}{{Model}} & Acc.@5 & BLEU@5 \\
\midrule
Look-Up Table (LUT) & 16.9 $\pm$ 0.2 & 25.2 $\pm$ 0.2 \\
T5-small (60M) & 37.8 $\pm$ 0.0 & 59.2 $\pm$ 0.5 \\
T5-large (770M) & 48.2 $\pm$ 0.0 & 69.1 $\pm$ 0.5  \\
T5-3B (3B) & 53.9 $\pm$ 0.0 & 72.3 $\pm$ 0.5 \\
\hline
BaseLLM$^*$ (best, $4shot$) & 43.0 $\pm$ 1.0 & 52.0 $\pm$ 1.4 \\
FT-LLM (no instr.) & \textbf{76.7 $\pm$ 1.1} & \textbf{83.9 $\pm$ 0.5} \\
FTnoise-LLM (no instr.) & 75.8 $\pm$ 0.7 & 83.4 $\pm$ 0.2 \\
\bottomrule
\end{tabular}
}
\caption{Comparing models (from Sec.~\ref{sec:experiments}) on the AE task on turn-2 given turn-1 as context. We report accuracy and BLEU score at top-5, as percentages, std. dev. computed on 3 runs. Higher is better, values in \textbf{bold} are highest in each column. The TDC-dev set was used for model selection before evaluation on test sets. 
\label{tab:tab_tdc_dev_table}}
\end{table}

\section{Computation cost}
\label{sec:computation_cost}

Fine-tuning of the 64B LLM uses TPU v3 with a 4x8 configuration, i.e., 32 TPUs. FT-LMM and  FTnoise-LLM are each trained for approximately 2100 and 1800 steps, respectively. The training time is approximately 3 hours. This leads to a model fine-tuning budget of 32 x 3 = 96 TPU * hour per model. 

Evaluation and inference on the 64B LLM uses TPU v3 with a 4x4 configuration, i.e., 16 TPUs. Each example (batch size = 128 samples) takes 0.653 s. This leads to $16\times0.568 / 128$ = 0.071 TPU $\times$ second per sample.

\section{Splitting characters in abbreviations.}
\label{sec:appx_char_split}
Pilot experiments showed the importance of programmatically inserting spaces between characters in the abbreviations. Since the vocabulary used by the LaMDA models is fairly large (32k entries), unless we enforce character-level splitting, subsequences of multiple characters in many abbreviations will be combined into spurious tokens, leading to slightly reduced AE accuracy.

\section{Recovery from failure - analysis}
\label{sec:appx_recovery_failure}
In the best scenario of replying to a question, the fine-tuned LLM is capable of predicting the correct phrase expansion approximately 81\% of the times with top-5 options and sufficient sampling (Figure \ref{fig:num_samples_acc_lat}). Hence the model will fail to find the correct expansion at least 19\% of the cases.

\section{Inference latencies of different LaMDA model sizes}
\label{sec:appx_latency}
In Figure~\ref{sec:appx_latency} we compare the latencies during inference time for the decoder-only models of different sizes. Compared to the 4B model, the 27B model shows 1.5x latency, while the 64B model shows 2.2x latency. While the latency increase is quite significant, this analysis shows that we cannot substitute the 64B model with a smaller model (e.g., by increasing the number of samples) in a way that improves latency without significantly harming the AE accuracy (compare the AE accuracies in Figure~\ref{fig:num_samples_acc_lat}.)

\begin{figure}
    \centering
    \includegraphics[width=\columnwidth]{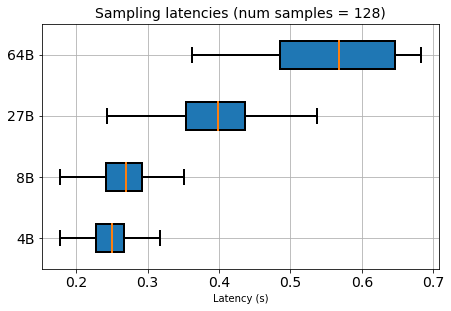}
    \caption{Inference latencies for different sizes of the LaMDA model (4B, 8B, 27B, and 64B.) The latencies are shown as box plots.}
    \label{fig:latency_comparison}
\end{figure}

\end{document}